\documentclass{article}

\usepackage[nonatbib, preprint]{tackling_climate_workshop_style}
\usepackage[utf8]{inputenc} 
\usepackage[T1]{fontenc}    
\usepackage{hyperref}       
\usepackage{url}            
\usepackage{booktabs}       
\usepackage{amsfonts}       
\usepackage{nicefrac}       
\usepackage{tabularray}
\usepackage{microtype}      
\usepackage{graphicx,float}

\newcommand\independent{\protect\mathpalette{\protect\independenT}{\perp}}
\def\independenT#1#2{\mathrel{\rlap{$#1#2$}\mkern2mu{#1#2}}}
\usepackage[
backend=biber,
style=numeric,
sorting=ynt
]{biblatex}
\usepackage{graphicx}
\graphicspath{
    {figures/}
}
\usepackage{subcaption}
\usepackage{rotating}
\usepackage[font=small,labelfont=bf]{caption}

\title{Causal Modeling of Soil Processes \\
for Improved Generalization} 

\author{%
  Somya Sharma  \\
  University of Minnesota Twin Cities\\
  \And
  Swati Sharma\thanks{Corresponding author. \texttt{swatisharma@microsoft.com} } \\
  Microsoft Research, Redmond  \\
  \AND
  Andy Neal \\
  Rothamsted Research, UK \\
  \And
  Sara Malvar \\
  Microsoft Research, Brazil \\
  \And
  Eduardo Rodrigues \\
  Microsoft Research, Brazil \\
  \AND
  John Crawford \\
  University of Glasgow, UK \\
  \And
  Emre Kiciman \\
  Microsoft Research, Redmond \\
  \And
  Ranveer Chandra \\
  Microsoft Research, Redmond \\
}

\begin{document}

\maketitle

\begin{abstract}
    Measuring and monitoring soil organic carbon is critical for agricultural productivity and for addressing critical environmental problems. Soil organic carbon not only enriches nutrition in soil, but also has a gamut of co-benefits such as improving water storage and limiting physical erosion. Despite a litany of work in soil organic carbon estimation, current approaches do not generalize well across soil conditions and management practices. 
    We empirically show that explicit modeling of cause-and-effect relationships among the soil processes improves the out-of-distribution generalizability of prediction models. We provide a comparative analysis of soil organic carbon estimation models where the skeleton is estimated using causal discovery methods. Our framework provide an average improvement of 81\% in test mean squared error and 52\% in test mean absolute error.
\end{abstract}

\section{Introduction}



Soil organic carbon, the carbon component of organic compounds found in soil both as biomass and as sequestered compounds and necromass,  has been called ``natural insurance against climate change'' \cite{droste2020soil}---with evidence associating increased soil organic matter with increased crop yields~\cite{SCHJONNING201835, soil-5-15-2019, Lal27A}. 
Climate change is increasing the variability in crop yields and increasing  food insecurity~\cite{kukal2018climate,ray2015climate, de2016towards}. This variability in yield is further exacerbated by conventional soil management practices unconcerned with soil organic carbon. 
Proper management of soil, including its organic carbon component, can mitigate shortages in food, water, energy and adverse repercussions of climate change~\cite{de2016towards}. 
Measuring and monitoring soil organic carbon can therefore have a positive impact in solving several environmental problems. This has led to increased interest by environmentalists, economists and soil scientists, as interdisciplinary collaborations, in improving public awareness and policy making \cite{lal2004managing, nziguheba2015soil, white2005principles, bhattacharyya2015soil, minasny2017soil, lal2004soil}. 

While the problem of studying soil organic carbon is well-motivated, forming hypotheses and designing experiments to estimate soil organic carbon can be challenging. Changes in soil organic carbon are not only dictated by weather events and management practices, but also by other soil processes such as plant nutrient uptake, soil organisms, soil texture, micro-nutrient content and soil disturbance. This makes soil a complex ``living'' porous medium \cite{de2014soil, de2016towards, ohlson2014soil}. 
Due to the complex nature of soil science, the exact relations among all soil processes is not yet known. There is no accepted universal method for studying soil organic carbon and the relation among soil processes \cite{bouma2002land}. Moreover, current models of soil organic carbon (RothC-26.3~\cite{rothc}, Century~\cite{century}, DNDC~\cite{GILTRAP2010292}, and some inter-model comparisons \cite{palosuo2012multi, tupek2019evaluating, todd2013causes, smith2012crop}) are not consistent with the latest advances in understanding of soil processes and do not generalize to different soil conditions found globally as they are limited by their modeling assumptions\cite{modelingsoilprocesses}. 

Our goal is to create a method that can generalize well across regions, soil types and soil management practices. We aim to create a causal machine learning framework that can aid in standardization efforts for soil organic carbon measuring, reporting and verification. 
 Although ML methods have demonstrated an improved ability to predict soil organic carbon  \cite{nguyen2021predicting, padarian2020machine}, the reliance of conventional ML on the i.i.d. assumption that training data represents the deployed environment limits their out-of-distribution generalizability \cite{ wadoux2020machine, padarian2020machine,  grunwald2022artificial, emadi2020predicting}. To improve this, either large (and diverse) data sets can be utilized or careful architectural modifications can help in creating a surrogate model or an emulator of the real-world physical system. These architectural modifications are also limited by domain experts' understanding of the physical system \cite{lavin2021simulation}. Using causal discovery methods is a way to overcome this limitation as these frameworks help explicitly model cause-and-effect relations among the soil processes to improve the out-of-distribution generalizability. 

In this paper, we present an approach based on causal graphs to estimate soil organic carbon stocks. We provide a comparative analysis of soil organic carbon estimation using causal and non-causal approaches and show that causal approaches produce better results for soil organic carbon estimation on unseen fields (from different locations, with different soil properties, management practices and land use).   
We briefly discuss related literature in Section~\ref{sec:relatedwork}, followed by the problem formulation, data set and our methodology in Section~\ref{sec:method}. Results, discussion and future work is discussed in Section~\ref{sec:experiment}.

\section{Background} 
\label{sec:relatedwork}
Our approach combines recent advances in causal discovery and graph neural networks (GNNs).  Causal discovery \cite{vowels2021d, glymour2019review, spirtes2016causal} is an approach for identifying cause-and-effect relations between variables of a system, using data under causal ignorability and sufficiency assumptions and leveraging partial {\em a priori} knowledge of relationships. 
Utilizing such causal discovery methods can quantify complex interactions of the different soil processes that govern soil organic carbon and its effects on soil quality, which cannot be directly measured but are emergent properties. Quantifying how soil organic carbon stocks are influenced by other soil process, and how soil organic carbon affects other soil processes and soil functions can move us closer to a universal or standardized modeling framework for all soil processes and for measuring soil organic carbon. Also, graph neural networks~\cite{zhang2019heterogeneous, zhou2020graph, chen2022identifying, tulczyjew2022graph} present approaches to work with non-Euclidean graph data and model complex relationships between entities. A survey of recent GNN advancements can be found here \cite{Wu2021GNN}.  Typically, GNN based methods assume homogeneity of nodes. Direct application of GNN approaches is not straightforward when nodes and edges are heterogeneous; this is the case in our application, where both the nature of the node (nodes could represent soil processes, climate variables, management practices) and the type of data associated with each node (e.g., soil process nodes could constitute continuous geochemical composition changes while farming or management practices might be frequency of an operation being performed on the farm) differ markedly. 

\section{Materials and Methods}
\label{sec:method}
\subsection{Data}
We utilize an extensive and rich data set from the North Wyke Farm Platform  \url{http://www.rothamsted.ac.uk/north-wyke-farm-platform}. This data is available for multiple fields with different land use types, which makes it appropriate for studying spatial out-of-distribution generalization, and limits bias due to causal ignorability and sufficiency assumptions.  The North Wyke Farm Platform data consists of observations of three pasture-based livestock farming systems, each consisting of five component catchments of approximately 21 ha each. High resolution long term data including soil organic carbon, soil total nitrogen, pH as well as management practices are collected to study the sustainability of different types of land use (treatments) over time (2012 to present).  In the baseline period (April 2011 to March 2013), all three farming systems were managed as permanent pastures, grazed by livestock and sheep. In April 2013, one system was resown with high sugar grasses, having a high water-soluble carbohydrate content with the aim of increasing livestock growth (``the red system''), a second system was resown with a high sugar grass-white clover mix (``the blue system'') to reduce the requirement for inorganic nitrogen fertilizer application. The remaining (``the green system'') continued as a permanent pasture for long-term monitoring.  Appendix~\ref{appendix:nwfp} shows a map of the North Wyke Farms showing the layout of the individual farms and their management practices~\cite{nwfp_map}.    

We create a train-test split to ascertain the generalizability of the proposed approach when fields are managed differently. For example, inversion ploughing is an important management practice because it results in the loss of organic carbon from agricultural soils\cite{Haddaway2017carbontillage}.  Figures~\ref{fig-soilcarbon} and~\ref{fig-plough} show the distribution of soil organic carbon and number of times the fields were ploughed for our data. Note that the red and blue systems were ploughed whereas the green system was not. This is also seen as a consistent higher levels of carbon for the green fields than the red and blue fields. Training data consists of 7 red and 8 blue system fields. Test data consists of a total of 7 green system fields. Our data set comprises management practices (including the number of fertilizer applications, pesticide applications, plough events, etc.), total nitrogen, total organic carbon and soil pH for each field. More details on data preprocessing are included in Appendix~\ref{appendix:data_preprocess}.

\subsection{Problem Formulation}
Complex interactions between soil organic carbon, soil processes and other exogenous factors (e.g., environmental and management practices) limit the generalization capabilities of conventional ML methods. 
In this paper, our aim is twofold, understanding how different management practices affect soil organic carbon and then  estimating it in a way that it generalizes in out-of-distribution environments. 
Let $M={M_1, M_2, M_3, \dots, M_n}$ represent farm management practices, $C$, $O$ represent soil organic carbon and other observed soil properties respectively, studied across $k$ locations.

Our approach is first to learn the cause-and-effect relations among soil variables (in this study they are, soil organic carbon, nitrogen and pH) and the management practices followed in the farm. The learned causal graph of different soil processes can be represented as $G \in \mathbb{R}^{{(|M| + |O| + |C|)} \times {(|M| + |O| + |C|)}}$. Depending on the causal discovery method employed for learning the causal graph, edge indices and attributes can be derived to create a skeleton that can be utilized in GNN-based regression to estimate soil organic carbon at a location as a function of the other variables. The generalization power of a causal graph skeleton based upon a GNN model relies on the graph $G$ that is used as prior knowledge for the prediction task. Here, for the regression task, instead of measuring the conditional expected response $\mathbb{E}(C|M, O)$,  we evaluate $\mathbb{E}(C|M, O, G)$  which is influenced by not only $p(D)$ but also causal graph $G$, where $D = \{ C, M, O \}$. Depending on the causal discovery method, additional assumptions can be made about the data \cite{spirtes2016causal}. 

\subsubsection{Causal Discovery}
In our experiments, a causal graph  consists of nodes---representing variables or physical processes and directed edges represent causal relationships between nodes. While prior knowledge or trial-and-error guessing can be used to create causal graphs, we make use of established causal discovery algorithms to create the directed graphs. To generate causal graphs using the North Wyke Farm Platform data, we use the PC algorithm, a  constraint-based method,  \cite{spirtes2000causation} and two score-based methods, Greedy Equivalence Search (GES) \cite{chickering2002optimal, meek1997graphical} and Greedy Interventional Equivalence Search (GIES) \cite{hauser2012characterization}. See Appendix~\ref{appendix:causal-discovery} for more details. 
\subsubsection{Causal Graph Neural Network}
While causal graphs estimated from causal discovery methods are used to obtain skeletons for GNNs, we compare two paradigms, Edge-Conditioned Convolution Message Passing Neural Networks (ECC MPNNs) \cite{simonovsky2017dynamic, ,gilmer2017neural} and GraphSAGE \cite{hamilton2017inductive}. Comparing different message passing procedures allows us to study how added complexity in learning influences generalization in the prediction task. These methods adopt different neighborhood definitions to compute message passing signals. For a directed graph $G (V, E)$, where $V$ is a finite set of nodes and $E$ is a set of edges, we can define a neighborhood for a given node $i$ as $N(i)$. For an ECC MPNN, $N(i)$ comprises all of  the ancestors in the directed graph. In GraphSAGE, a neighborhood is defined as a function of varying search depths $k \in \{0, 1, ..., K \}$, wherein, the number of adjacent nodes are sub-sampled for message passing with a node $i$. Starting at $k=0$, the neighboring feature vectors are aggregated at each search depth $k$ and concatenated with a node's representation. The final representation is obtained as the aggregation at depth $K$. More details on how node embeddings are updated are mentioned in Appendix~\ref{appendix:gnn-approaches}. 

\section{Results and Discussion}
\label{sec:experiment}

\begin{figure}
    \centering
    \begin{minipage}{0.35\textwidth}
        \centering
        \includegraphics[width=\textwidth]{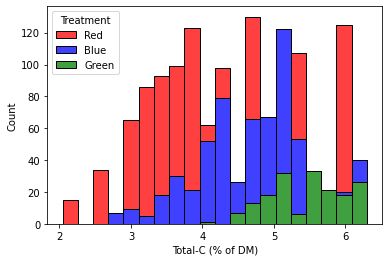}
\caption{Soil organic carbon distribution}
\label{fig-soilcarbon}

\includegraphics[width=\textwidth]{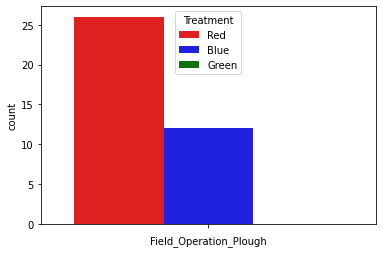}
\caption{Plough event distribution for the three treatments}
\label{fig-plough}

    \end{minipage}\hfill
    \begin{minipage}{0.6\textwidth}
        \centering
        \small
        \begin{tblr}{r|c|c|c}
          \hline
          {\bf Model} & {\bf Causal} & {\bf MSE} & {\bf MAE}  \\
          \hline
         PC + GraphSAGE & Yes & 1.1002 & 1.0116  \\  
         GES + GraphSAGE & Yes & 0.9248 &  0.9193 \\
         GIES + GraphSAGE & Yes & 0.7803 &  0.7904 \\
         PC + ECC MPNN & Yes & \textbf{0.2816}  & \textbf{0.5258}  \\
         GES + ECC MPNN & Yes & 0.2864 & 0.5302 \\
         GIES + ECC MPNN & Yes & 0.2951  & 0.5385  \\         \hline[dashed]
         Random Edges + GraphSAGE & No & 2.9052  & 1.6686   \\
         XGBoost & No & 4.5007 & 2.0860 \\

         MLP & No & 3.8415 & 1.9254 \\
         Random Forest & No & 2.7996 & 1.6263 \\
         \hline
         
    \end{tblr}
    \captionof{table}{Comparison of soil organic carbon estimation approaches based on Mean Squared Error (MSE) and Mean Absolute Error (MAE) to show how well different approaches generalize for permanent pasture i.e. the ``green system''. We use high sugar grass pastures (``red'' and ``blue'' systems) for training and the ``green" system for testing. We compare GraphSAGE architecture and ECC MPNN architecture.} 
    \label{tab:performance_table_v2}
    \end{minipage}
\end{figure}

Our experiments investigate the impact of soil processes and farming practices on soil organic carbon estimation. Through empirical evidence, we demonstrate the improvement in out-of-distribution generalization offered by coupling causal discovery methods with GNNs. 
Results in Table~\ref{tab:performance_table_v2} suggest that causal approaches outperform non-causal approaches for soil organic carbon estimation and generalize well to unseen locations. 
The causal graph generated by the PC method is more parsimonious than the score-based methods' graphs and 
offers the best prediction skill when used as skeleton for ECC MPNN model. The causal graphs resulting from the 3 causal discovery approaches are in Appendix ~\ref{appendix:causal-graphs} and 
The details of ML algorithms and hyperparameter tuning are in Appendix~\ref{appendix:hyper}. 




Machine learning models learn parameters that fit the training data. This is problematic if the training data is constrained, for example, in this study the training data only consists of fields that were tilled. Soil carbon change due to tilling is more pronounced. If test data comes from the same data distribution as we trained on, the machine learning model will perform well because the same learned parameters from the training data apply. 
When we test on out-of-distribution data (i.e., no tilling), the same learned parameters do not apply anymore; consequently the machine learning models do worse. 
Causal approaches generalize better because they guide the models towards the true underlying real-world process. Due to this reason, causal approaches are well-suited to solve a major drawback of the existing soil models: they have to be re-calibrated and they do not work well for different types of soil types and management practices. Using causal discovery approaches allows for automatically discovering the ``true'' links between different soil factors and management practices so we know how the soil dynamics are affected when certain management practices are used. We discuss some uncovered relationships in Appendix~\ref{appendix:causal-graphs}. We are currently doing a large scale study to include more soil types and management practices. Finally, the presented causal approaches have broader applicability such as modeling gene-environment interaction where we require generalization to new environments, such as new locations with different weather patterns 
 or new seed types. 

 

\small
\printbibliography

\normalsize

\appendix
\section{Appendix}
\subsection{Data preprocessing}
\label{appendix:data_preprocess}
Field names (22 fields) and management practices (54 practices) are one hot encoded. Numerical data (i.e., total Nitrogen, total carbon and soil pH) is scaled using a min-max scaling scheme. In addition, to understand the long-term cumulative effects of changes in management practices, we include lag variables that capture the number of times a management practice / farming operation was performed in the last 1.5 months, 6 months, 1 year, 2 years. 
Different data are available at varying cadence, so we merge them together at a daily level by averaging the values of features collected at a finer resolution. 
\subsection{North Wyke Data Farm}
\label{appendix:nwfp}
Layout of the North Wyke Data Farm showing color coded fields to represent the land use type. The red system is sown with high sugar grasses, having a high water-soluble carbohydrate content with the aim of increasing livestock growth, the blue system sown with a high sugar grass-white clover mix to reduce the requirement for inorganic nitrogen fertilizer application. The remaining fields continued as a permanent pasture for long-term monitoring (the green system).
\begin{figure*}[h]
\centering
\includegraphics[height=0.6\textheight]{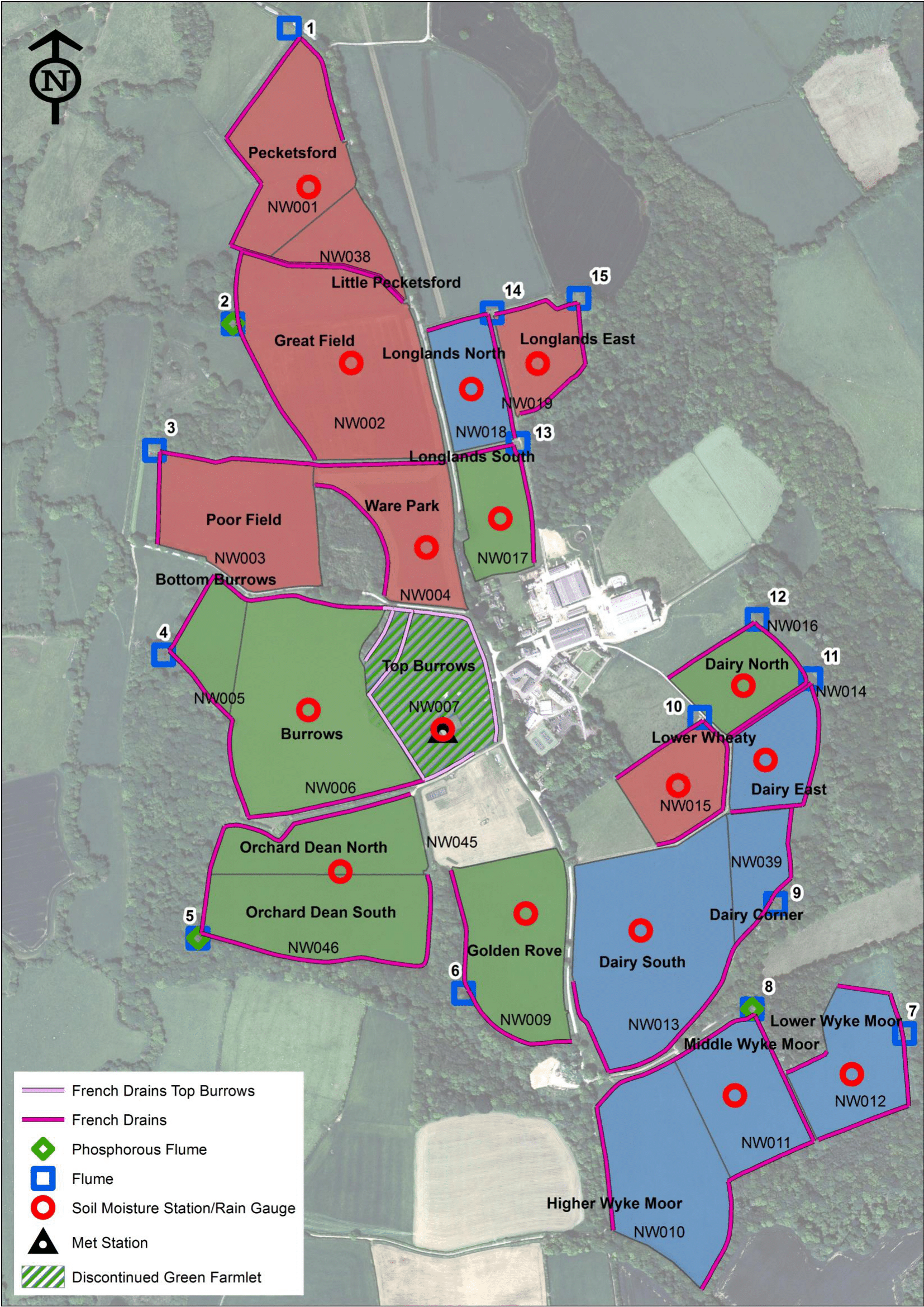}
\caption{Layout of the North Wyke Farms}
\label{fig:nwfp_map}
\end{figure*}

\subsection{Causal discovery approaches: PC, GES and GIES}
\label{appendix:causal-discovery}
The PC algorithm relies on conditional independence testing to establish causal relations. An undirected graph is used as an initial skeleton and edges between independent variables are eliminated. Conditional independence of variables conditioned on a set $S$, $D_i \independent D_j | S$, is evaluated to eliminate additional edges \cite{vowels2021d}. The score-based methods evaluate the fitness of causal graphs based on a scoring function to obtain an optimal graph \cite{peters2017elements}. In the case of GES, the Bayesian Information Criterion (BIC) is used as the scoring function. Starting with an empty graph, edges are added if they improve (lower) the score. The graph is then mapped to a corresponding Markovian equivalence class followed by elimination of edges that may provide further improvement, assessed using the BIC \cite{glymour2019review}. Similar to GES, GIES also utilizes a quasi-Bayesian score and searches for the causal graph that optimizes for the score. GIES is a generalization of GES that incorporates interventional data. Apart from adding edges (forward phase) and removing edges (backward phase) that improve score, GIES introduces a ``turning phase" to improve estimation wherein at each iteration, an edge is turned to obtain an essential graph with same number of edges.

\subsection{Details of GNN approaches}
\label{appendix:gnn-approaches}

\begin{figure}[h]
\begin{center}
   \includegraphics[width=0.5\textwidth]{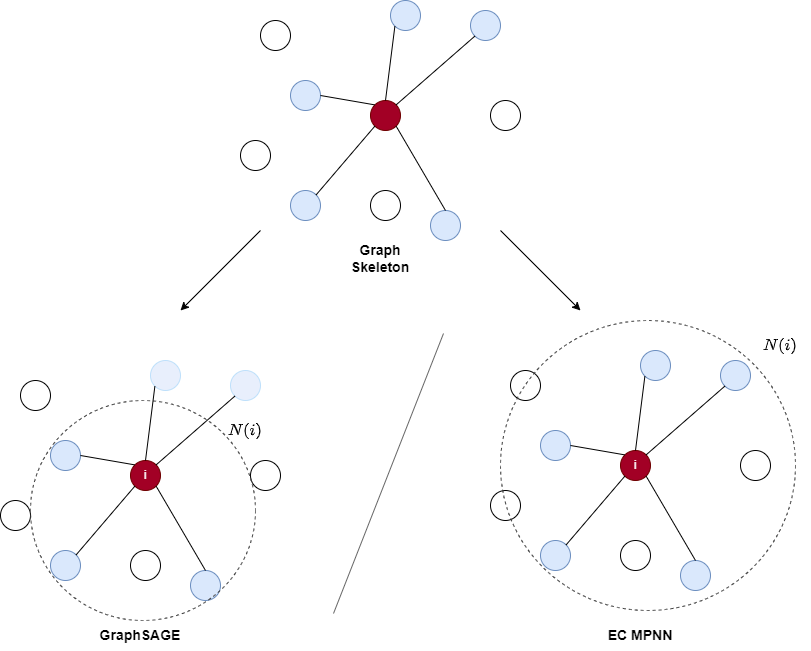}
 \end{center}
  \caption{Neighborhood definition in Causal Graph Neural Networks, GraphSAGE and ECC MPNN. ECC MPNN defines all connected adjacent nodes as neighborhood while GraphSAGE aggregates information over different depths, starting from depth 0 (node features itself used as neighboring feature vector) till depth $K$, where, at each depth $k$, connected nodes as sub-sampled to be included in neighboring feature vector.} 
\label{fig: cgnn-neighborhood}  
\end{figure}

For ECC MPNN, at each layer $l$ of the feed-forward neural network, the embedding signal can be computed as,

\begin{equation}
    \mathbf{h}^l_i \leftarrow \frac{1}{|N(i)|} \sum_{j \in N(i)} F^l (E_{j, i}; W^l)h^{l-1}_j + b^l
\end{equation}

where, $W^l$ and $b^l$ are the weight matrix and the bias term defined at layer $l$. In GraphSAGE, embeddings at search depth $k$ for given node $i$ can be computer as,

\begin{equation}
    \mathbf{h}^k_i \leftarrow \sigma ( W^k [\mathbf{h}^{k-1}_i, AGG(\{ \mathbf{h}^{k-1}_{u}, \forall u\in N(i)\})] )
\end{equation}

where, $\sigma$ is a non-linear activation function and $W^k$ is the weight matrix at depth $k$. $\mathbf{h}^k_{N(i)} = AGG(\{ \mathbf{h}^{k-1}_{u}, \forall u\in N(i)\})$ is the signal aggregated over all sub-sampled neighbors at depth $k$. $AGG$ can be any aggregator function including trainable neural network aggregator. Architectures for both methods include paradigm based convolution layer followed by linear layers and then ReLU activation. Figure~\ref{fig: cgnn-neighborhood} shows the neighborhood definition for the two GNN approaches considered. 

\subsection{Architectural choices and hyperparameter tuning}
\label{appendix:hyper}
In the GraphSAGE architecture, we choose $K$ to be number of connected nodes, mean as the aggregator function, $AGG$. In ECC MPNN network, since we have only one type of edge feature in the form of directed edge existence, number of edge feature is set to 1. Model hyperparemeters are chosen from grid search - Adam optimization method for both GNNs, learning rate 0.0015 for GraphSAGE and 0.0020 for EC MPNN. The GraphSAGE architecture uses three sequential GraphSAGE convolution layers to learn embeddings. The architecture also comprises three feed-forward layers, used to generate estimates for the target variable from the embeddings. For the ECC MPNN model, two ECC convolution layers are stacked sequentially  to estimate the embeddings. A final feed-forward layer is then used to learn the target variable estimates. We combine three causal discovery methods, PC, GES and GIES, with the two GNN architectures to obtain 6 variants of Causal GNNs. We compare these with four benchmarks, XGBoost (100 estimators, 20 max depth), Random Forest (100 estimators), MLP (random grid search based hyperparameter set) and Random Edges + SageGRAPH (50 random directed edges used as skeleton).

\subsection{Causal graphs}
\label{appendix:causal-graphs}
In this section, we present the causal graphs produced by the three algorithms: PC algorithm \cite{spirtes2000causation} Greedy Equivalence Search (GES) \cite{chickering2002optimal} and Greedy Interventional Equivalence Search (GIES) \cite{hauser2012characterization}. The nodes of the graph represent the features considered that include one hot encoded fields (nodes named as Field\_\textit{field\_name}), one hot encoded management practices (Field\_Operation\_\textit{operation}), total Nitrogen (total-N), total carbon (total-C) and soil pH (pH). The existence of an edge represents a causal relationship and the direction of the arrow represents the direction of influence. In all the three graphs, we notice that there is a strong association between Total Carbon (\% of DM) and Total Nitrogen (\% of DM) in soil and soil organic matter (SOM). These relationships are consistent with theory and other studies~\cite{CtoN}. Also, note that fertilizer addition and manure addition adds more total Nitrogen~\cite{MENG20052037}. Other well known relationships we discover are that ploughing changes the pH of the soil, which affects total carbon~\cite{tillage_ph}.  
\begin{sidewaysfigure}[h]
\includegraphics[width=\textwidth]{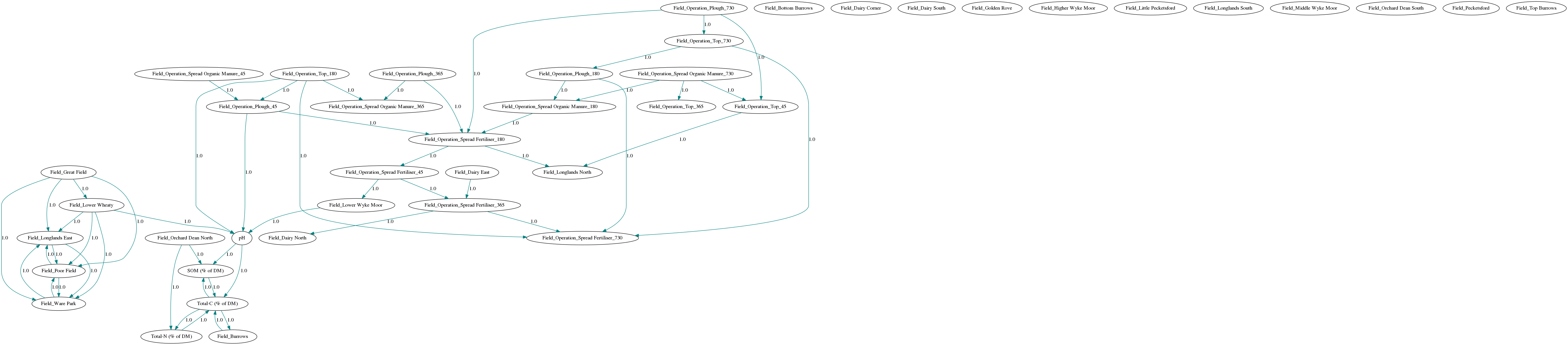}
\label{fig:causalgraph-pc}
\caption{Causal graph generated by PC algorithm}
\end{sidewaysfigure}

\begin{sidewaysfigure}[h]
\includegraphics[width=\textwidth]{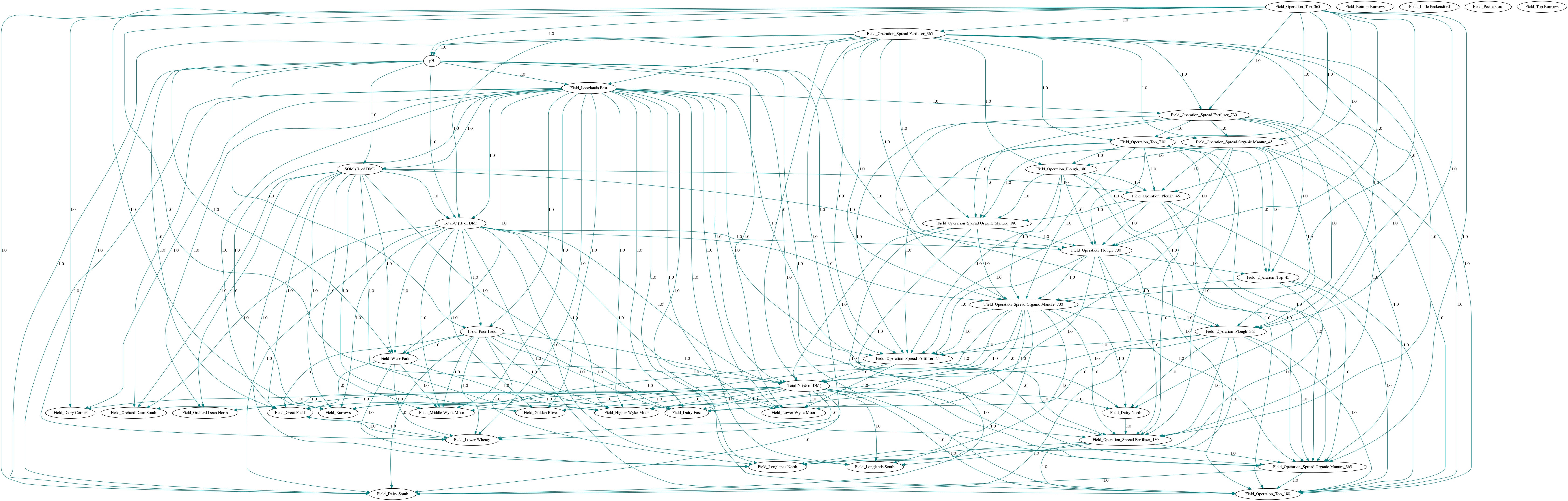}
\caption{Causal graph generated by GES algorithm}
\label{fig:causalgraph-ges}
\end{sidewaysfigure}

\begin{sidewaysfigure}[h]
\includegraphics[ width=\textwidth]{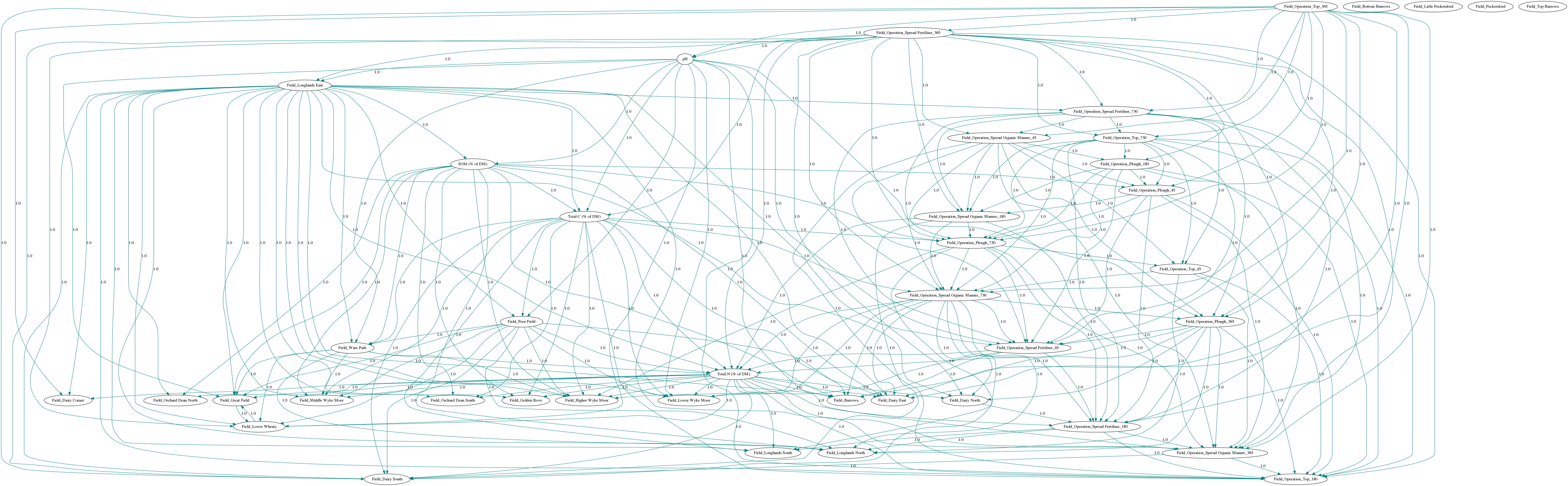}
\label{fig:causalgraph-gies}
\caption{Causal graph generated by GIES algorithm}
\end{sidewaysfigure}

\end{document}